\pdfoutput=1

\documentclass[11pt]{article}

\usepackage[]{acl}

\usepackage{times}
\usepackage{latexsym}

\usepackage[T1]{fontenc}

\usepackage[utf8]{inputenc}

\usepackage{microtype}

\usepackage{graphicx}
\usepackage{xparse}

\NewDocumentCommand\emojienvelope{}{
    \includegraphics[scale=0.035]{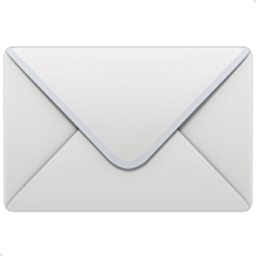}
}
\NewDocumentCommand\emojilaptop{}{
    \includegraphics[scale=0.035]{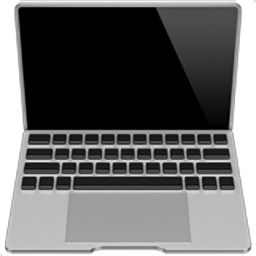}
}

\title{The Role of Summarization in Generative Agents: \\ A Preliminary Perspective}

\author{
Xiachong Feng, 
Xiaocheng Feng,
Bing Qin \\
Harbin Institute of Technology, China\\
\texttt{\{xiachongfeng,xcfeng,bqin\}@ir.hit.edu.cn}
}

\begin{document}
\maketitle
\begin{abstract}
Generative agents \cite{park2023generative} that simulate human society show tremendous potential for further research and practical applications.
Specifically, the generative agent architecture comprising several meticulously designed modules constitutes the most critical component.
To facilitate progress in this research, this report presents our integrated perspective on comprehending generative agents through summarization, since we believe summarization is the most fundamental and indispensable capacity of generative agents manifested across diverse scenarios.
We hope this report can provide insight into understanding the importance of summarization capacity in generative agents and motivate future research.
\end{abstract}

\section{Introduction}

Recent advancements in Large Language Models (LLMs), such as ChatGPT and GPT-4 \cite{OpenAI2023GPT4TR}, have rebuilt various domains including natural language processing \cite{Yang2023HarnessingTP}, computer vision \cite{Wu2023VisualCT} and autonomous robotics \cite{Mai2023LLMAA}. These cutting-edge models enable novel opportunities to achieve artificial general intelligence (AGI).
Owing to the rapid progress of LLMs, there is an emerging consensus that LLMs have attained preliminary intelligence and now demonstrate comparable performance to humans on various tasks \cite{Zhao2023ASO}.

In the current era of large language models, \citet{park2023generative} propose Generative Agents: sophisticated computational software powered by fundamental language models that can simulate believable human behaviour within meticulously designed environments and protocols.
This well-designed framework offers comprehensive opportunities for exploring and understanding human social dynamics, including long-term goal planning, information transformation, relationship establishment and coordination.

In this report, we present our view on generative agents from the perspective of automatic summarization and demonstrate how various functional components of such agents can be formalized as summarization tasks.
Specifically, we identify several key summarization techniques that are integral to implementing generative agents: 
(1) The retrieve module contains the idea of unsupervised summarization (\S\ref{sec:us}); 
(2) The reflection module is composed of two sub-modules: extreme summarization (\S\ref{sec:es}) and citation-based summarization (\S\ref{sec:cs});
(3) Query-based summarization (\S\ref{sec:qs}) supports following Plan module and Act module;
(4) Summarization with emojis (\S\ref{sec:se}) provides an intuitive visual interface;
(5) The agent's movement in the environment can be abstracted to Graph Summarization (\S\ref{sec:gs});
and (6) Dialogue between agents is facilitated by Dialogue Summarization (\S\ref{sec:ds}).
We hope this paper illuminates the potential of summarization techniques in advancing the development of future generative agents.

\begin{figure*}[t]
    \centering
    \includegraphics[scale=0.32]{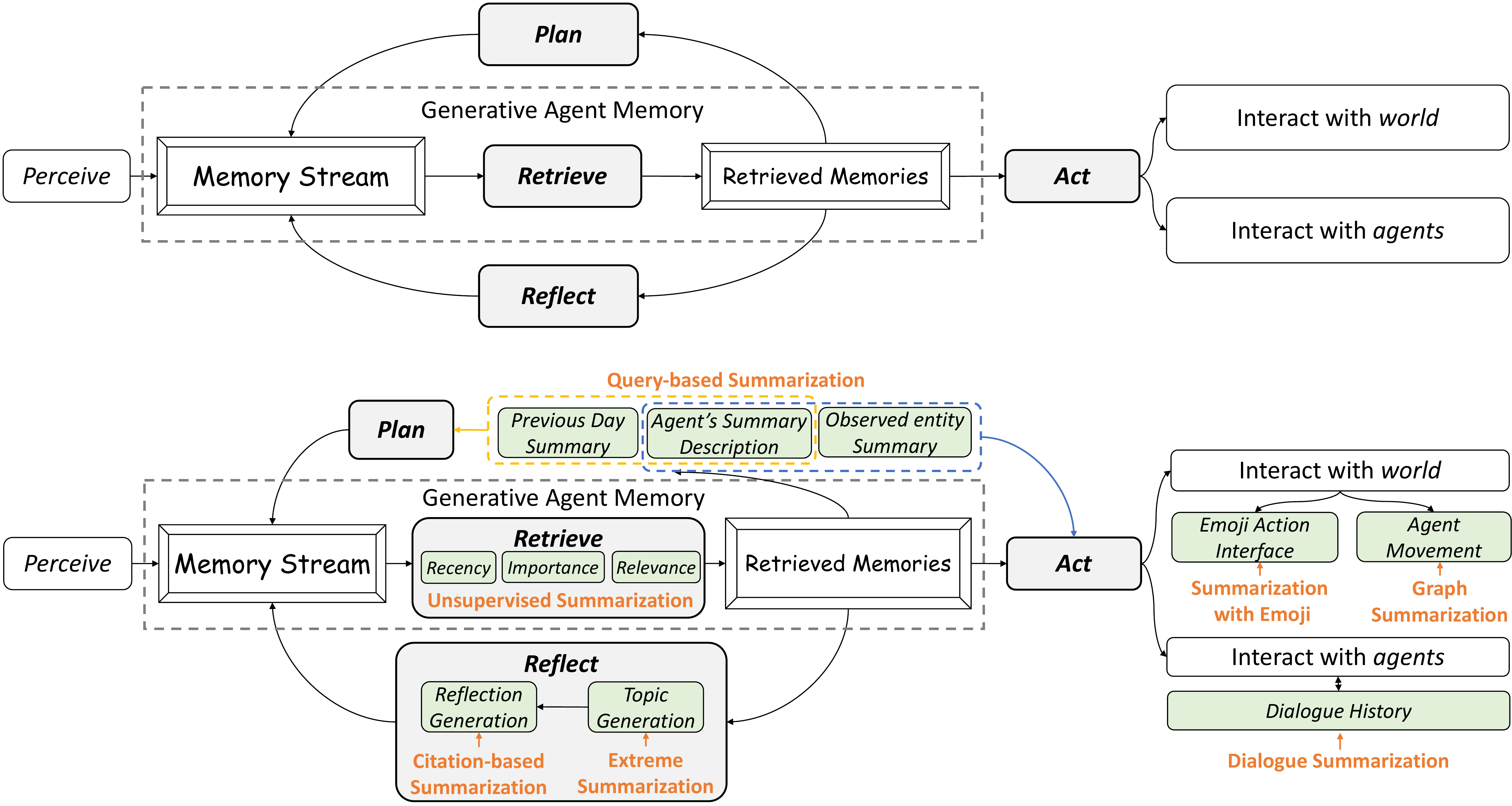}
    \caption{Illustration of the generative agent architecture and key summarization techniques inside the architecture.}
    \label{fig:key}
\end{figure*}

\section{Generative Agents}

Generative agents are AI-powered computational software that can simulate believable human behaviour.
In this section, we provide a concise overview of the generative agent architecture including several main components.
The overall architecture is shown in Figure \ref{fig:key}.
The fundamental \textbf{Memory} module is responsible for storing various types of information related to the agent itself, including basic \textit{observations} as well as high-level \textit{reflections} and generated \textit{plans}.
The \textbf{Retrieve} module then extracts appropriate memories from the memory stream to support downstream modules including \textit{Plan}, \textit{Reflect} and \textit{Act}.
Afterwards, the \textbf{Reflect} module provides high-level abstractions of one agent's memories, which serve as another type of memory.
Furthermore, the \textbf{Plan} module takes the agent's summary and the observed entity's summary into consideration and creates the plan in a course-to-fine manner.
Finally, agents \textbf{Act} with the world by performing actions or with other agents by initiating dialogues.

\section{Key Summarization Techniques}

\subsection{Unsupervised Summarization}\label{sec:us}
The \textbf{Retrieve} module aims to offer pilot memories given the agent’s current situation and the entire memory stream. 
This coincides with the objective of unsupervised summarization which seeks to extract the essential information from a collection of documents given one desired query based on various manufactured features \cite{Carbonell1998TheUO}.

Specifically, the retrieve function takes three distinct features, \textit{Recency}, \textit{Importance}, and \textit{Relevance}, into consideration to effectively derive prominent information from the memory stream. 
In detail, \textit{Recency} posits that recently accessed memories are important since we human beings are frequently processing short-term tasks.
\textit{Importance} is inferred directly by the LLM depending on its tremendous background knowledge by simply prompting the LLM.
\textit{Relevance} assigns a higher score to those most relevant memories with respect to the agent's current situation.
The final retrieval score is a weighted sum of three scores: $\mathrm{score}=\alpha_{\mathrm{recency}} \cdot \mathrm{recency} + \alpha_{\mathrm{importance}} \cdot \mathrm{importance} + \alpha_{\mathrm{relevance}} \cdot \mathrm{relevance}$.
With the integration of the above three features, the retrieve module successfully conducts unsupervised summarization over the memory to produce digest information for the following steps.

\subsection{Extreme Summarization}\label{sec:es}
\textbf{Reflection} is one of the most critical components of the generative agent, which summarizes the agent's recent situation and creates high-level thoughts.
The whole reflection can be divided into two steps, in which the first step is extreme summarization.
Concretely speaking, it aims to condense the agent's 100 most recent memory records into three key topics in the question generation manner.

Specifically, the module achieves the goal by prompting the LLM via ``Given only the information above, what are 3 most salient high-level questions we can answer about the subjects in the statements?".
The results include three highly condensed questions that can be viewed as extreme summaries of the agent's recent memories since several previous studies verify the tight connection between summarization and question generation \cite{Narayan2020QURIOUSQG,ijcai2021p524}.

\subsection{Citation-based Summarization}\label{sec:cs}
The second step of \textbf{Reflection} can be viewed as a citation-based summarization task, which receives several retrieved documents (memories) with indexes and aims to produce summaries with evidence references. This is also in line with the previous related work generation task and the open-domain reading comprehension task, both of which require providing concrete evidence to support their generated results \cite{chen-etal-2021-capturing}.

Specifically, the module achieves the goal by prompting the LLM via ``Statements about Klaus Mueller, ..., What 5 high-level insights can you infer from the above statements? (example format: insight (because of 1, 5, 3))". The output abstracts relevant memories into the reflection with citations: ``Klaus Mueller is dedicated to his research on gentrification (because of 1, 2, 8, 15)".

\subsection{Query-based Summarization}\label{sec:qs}
In fact, query-based summarization permeates the core architecture of the entire generative agent with the help of the \textbf{Retrieve} module.
In this part, we mainly focus on three tasks that will support the subsequent \textbf{Plan} and \textbf{Act} modules.

\paragraph{Agent's Summary Description}
Agent's summary description summarizes the agent's identity information, current occupation situation and self-assessment, which serves as a critical clue to making plans and taking reactions.
In detail, relevant memories are first obtained via three queries ``[name]’s core characteristics", ``[name]’s current daily occupation", and ``[name’s] feeling about his recent progress in life", and then three resulting summaries are combined into the whole agent's summary description.

\paragraph{Previous Day Summary}
Previous Day Summary plays an important role in the plan creation process, which ensures the agent achieves consistent and long-term goals.
Although no detailed information is provided in the original paper \cite{park2023generative}, we assume the implicit query ``[name]’s previous day plan" is used to retrieve relevant memories and produce the final summary.

\paragraph{Observed Entity Summary}
The observed entity summary that compresses (1) the relationship between the agent and the entity and (2) the status of the entity is an important basis for whether the agent takes action.
The summary consists of two parts obtained via queries ``What is [observer]’s relationship with the [observed entity]?” and ``[Observed entity] is [action status of the observed entity]”.
Taking both agent's summary description and observed entity summary into consideration, the agent decides whether or not to react by prompting the LLM ``Should John react to the observation, and if so, what would be an appropriate reaction?" 

\subsection{Summarization with Emojis}\label{sec:se}
To give quick access to the agent's status, \citet{park2023generative} implements a high-level emoji-based abstraction on the sandbox interface by prompting the LLM.
For example, ``Isabella Rodriguez is checking her emails" appears as \emojilaptop\emojienvelope.
As the saying goes, a picture is worth a thousand words, the emoji interface intuitively summarizes the agent's current status and integrates into the whole system.

\subsection{Graph Summarization}\label{sec:gs}
Agents who lived in Smallville can perform movements to reach the appropriate location.
The Smallville realizes a tree representation, where the root node denotes the entire world, children nodes describe areas and leaf nodes indicate objects.
The agent's movement is decided by first transforming the tree representation into natural language and then prompting the LLM via ``Which area should [name] go to?".
In other words, the movement of an agent can be formalized as an implicit graph summarization task \cite{kaushik2003graph}.
Given the world graph, the agent finds one suitable path from the current location towards the target destination.

\subsection{Dialogue Summarization}\label{sec:ds}
The agents interact with each other through dialogue.
At the initial point, one agent decides to trigger the dialogue based on the action given the agent’s summary description and observed agent (entity) summary.
To make the dialogue coherent and informative, the following utterances are generated by considering additional dialogue summaries. 
In the original paper \cite{park2023generative}, pure dialogue histories are used to facilitate dialogue generation.
We believe that when facing long and verbose dialogue histories, dialogue summarization can be an effective method to address such a challenge.
Additionally, on the demo page, the dialogue summary also provides a quick overview of the core contents of a dialogue.

\section{Conclusion}
In this report, we aim to understand generative agents from a unified view of summarization.
We systematically analyze several key summarization techniques and show how individual modules inside the generative agent architecture can be formalized as traditional summarization tasks.
We believe future generative agents can be substantially enhanced with advanced summarization abilities.

\bibliography{custom}
\bibliographystyle{acl_natbib}

\end{document}